\documentclass{article}

\usepackage{amsfonts}

\usepackage{microtype}
\usepackage{graphicx}
\usepackage{subfigure}
\usepackage{booktabs} 

\usepackage{hyperref}

\usepackage[accepted]{icml2021}

\icmltitlerunning{Evaluating the Use of Reconstruction Error for Novelty Localization}

\usepackage{xcolor}

\usepackage{titlesec}
\titlespacing\section{0pt}{0pt plus 2pt minus 0pt}{-2pt plus 2pt minus 0pt}
\titlespacing\subsection{0pt}{0pt plus 2pt minus 0pt}{-2pt plus 2pt minus 0pt}

\begin{document}

\setlength{\abovedisplayskip}{2pt plus 3pt}
\setlength{\belowdisplayskip}{2pt plus 3pt}

\twocolumn[
\icmltitle{Evaluating the Use of Reconstruction Error for Novelty Localization}



\icmlsetsymbol{equal}{*}

\begin{icmlauthorlist}
\icmlauthor{Patrick Feeney}{tufts}
\icmlauthor{Michael C. Hughes}{tufts}
\end{icmlauthorlist}

\icmlaffiliation{tufts}{Department of Computer Science, Tufts University, Medford, Massachusetts, USA}

\icmlcorrespondingauthor{Patrick Feeney}{patrick.feeney@tufts.edu}

\icmlkeywords{Machine Learning, ICML}

\vskip 0.3in
]



\printAffiliationsAndNotice{}  

\begin{abstract}
The pixelwise reconstruction error of deep autoencoders is often utilized for image novelty detection and localization under the assumption that pixels with high error indicate which parts of the input image are unfamiliar and therefore likely to be novel.
This assumed correlation between pixels with high reconstruction error and novel regions of input images has not been verified and may limit the accuracy of these methods.
In this paper we utilize saliency maps to evaluate whether this correlation exists.
Saliency maps reveal directly how much a change in each input pixel would affect reconstruction loss, while each pixel's reconstruction error may be attributed to many input pixels when layers are fully connected.
We compare saliency maps to reconstruction error maps via qualitative visualizations as well as quantitative correspondence between the top K elements of the maps for both novel and normal images.
Our results indicate that reconstruction error maps do not closely correlate with the importance of pixels in the input images, making them insufficient for novelty localization.

\end{abstract}

\section{Introduction}
\label{intro}

\begin{figure}[h!]
  \centering
  \includegraphics[width=\linewidth]{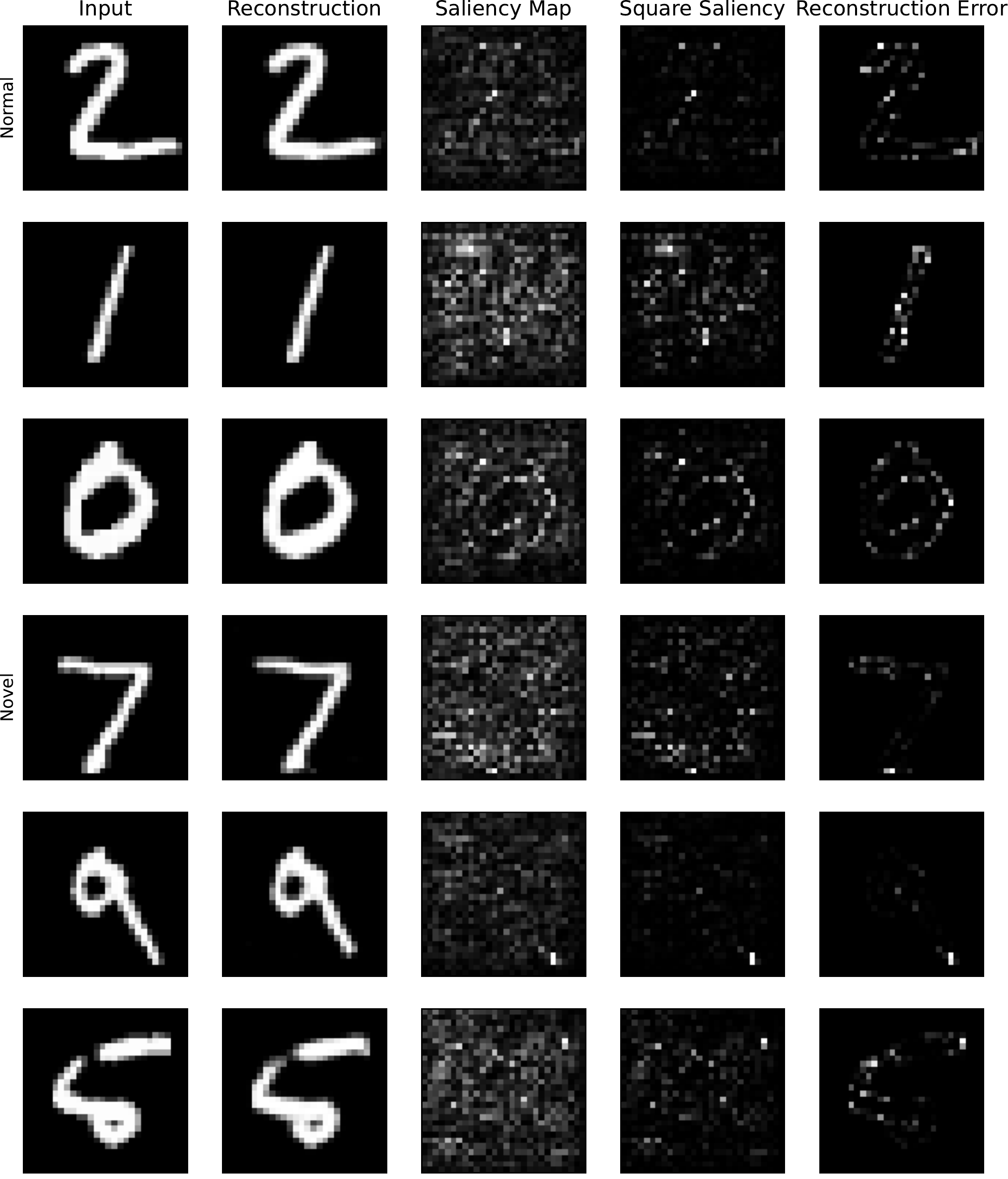}
  \vspace{-12pt} 
  \caption{Comparison of normalized maps produced by a deep autoencoder given 3 examples of ``normal'' input (\emph{top 3 rows}, digits 0-4) and 3 examples of ``novel'' input (\emph{bottom}, digits 5-9). From left to right: model input, input reconstructions, saliency map for reconstruction error, squared saliency map, and reconstruction error map. Saliency maps represent input pixels with a high impact on the reconstruction error. While reconstruction error maps share \emph{some} high intensity pixels with saliency maps, reconstruction error maps often fail to identify important input pixels highlighted in the saliency map that are \emph{far from the digit boundary}. This deficiency is more pronounced on novel inputs, thus calling into question the suitability of reconstruction error maps for novelty localization.}
  \label{fig:visual_comparison_saliency_vs_reconstruction}
\end{figure}

One-class image novelty detection is a classification task where a model must determine whether an image is normal, belonging to the same distribution as the training data, or is novel, belonging to a different distribution \cite{japkowicz1995novelty, abati19, perera19}. Unlike typical classification tasks, one-class novelty detection requires training only on data from a single class, often under the assumption that novel data could not be collected in an applied setting. This task requires distinct methods from standard binary classification, because models must be trained without examples of novel images. \citet{geng2020recent} contrasts one-class novelty detection with similar tasks such as open set recognition \cite{Oza_2019_CVPR} and out-of-distribution detection \cite{ren2019likelihood, liang2017enhancing}, noting that this task focuses solely on outlier detection instead of also classifying normal images.

Existing works often utilize deep autoencoders for image novelty detection. These models utilize the reconstruction error, having found that novel images will typically have a larger reconstruction error than normal images \cite{japkowicz1995novelty, richter2017safe}. Various models utilize this finding in different ways. Constructing an empirical cumulative distribution function for the reconstruction error on the training data and classifying images based on the reconstruction error percentile has been used for novelty detection in visual navigation tasks \cite{richter2017safe}. Some methods have increased the reconstruction error percentile of novel images by using generative adversarial networks or recurrent neural networks \cite{perera19, sabokrou2017deepanomaly, luo2017revisit}. The reconstruction error signal has also been combined with signals from a latent space representation learned via variational autoencoders~\cite{abati19} or a generative adversarial network~\cite{perera19}.

Once a novelty has been detected, some models attempt to \emph{localize} the novelty. The task of \emph{novelty localization} entails determining which pixels in the input image lead to the model labeling it as novel. Localization can be useful in practice for identifying an abnormal region in a medical image or a defective part of a manufactured object ~\cite{zimmerer2018contextencoding, Bergmann_2020_CVPR}. When utilizing a deep autoencoder, novelty localization is typically done using the \textit{reconstruction error map}, an image containing the per pixel reconstruction error produced by the autoencoder~\cite{abati19, sabokrou2017deepanomaly,Gong_2019_ICCV,Bergmann_2020_CVPR, zimmerer2018contextencoding}. Using this map assumes that pixels with high reconstruction errors correspond to novel pixels in the input image. Surprisingly, this assumption about the correlation between reconstruction error and input images has not been verified. Relying on this assumption may limit the accuracy of these methods.

We note the similarity between the novelty localization and classification model explanation problems. Both problems require the identification of input pixels with the largest influence on some image-level output function, novelty detection and classification respectively. In this paper we utilize saliency maps to evaluate whether reconstruction error maps correlate with the importance of pixels in the input images for deep autoencoder architectures. To assess this correlation between saliency maps and reconstruction error maps, we use both qualitative visualizations (see Fig.~\ref{fig:visual_comparison_saliency_vs_reconstruction}) and quantitative metrics. We introduce two metrics to better summarize the correspondence between two maps. Thorough experiments on MNIST digit data show the utility of our saliency map approach and metrics. Our results indicate that reconstruction error maps do not closely correlate with the importance of pixels in the input images, making them insufficient for novelty localization.

\section{Related Work}

\subsection{Deep Autoencoders for Novelty Detection}
\label{sec:rel_work_autoencoders_for_novelty}

The simplest deep autoencoder approaches to novelty detection consist of two parts: a deep autoencoder~\citep{bourlard88,vincent08} that produces a reconstruction map, and a method that consumes the reconstruction map and produces a decision about novelty. In the case of single image inputs, a threshold on the reconstruction error is often used to determine novelty \cite{japkowicz1995novelty,richter2017safe}. This threshold may be set using any method for hyperparameter selection, but thresholding at a high percentile of the empirical cumulative distribution function for the reconstruction error on the training data provides a low-cost solution \cite{richter2017safe}.

Recent advances tend to focus on introducing new methods for one part of the larger model. Autoencoder focused methods have created models that increase the gap between the reconstruction error of normal and novel images by using generative adversarial networks (GANs) during training \citep{perera19, sabokrou2017deepanomaly}. Adversarial training forces the network to decode more latent vectors into plausible, normal images. 
Other efforts use sparse coding representations within neural networks to similar effect~\citep{luo2017revisit}. Furthermore, latent representations can be constrained to a subset of the embeddings of training inputs \citep{Gong_2019_ICCV}.

Approaches focused on improving the novelty detection part introduce ways to utilize the latent space for detection. The probability of an input image being mapped to a latent vector can be probabilistically modeled via a Gaussian mixture model \citep{zong2018deep}, a Gaussian classifier \citep{sabokrou2017deepanomaly}, a variational autoencoder \citep{zimmerer2018contextencoding}, or an autoregressive process \citep{abati19}. This probability can be utilized independently or combined with reconstruction error to form a novelty score.

\subsection{Saliency Maps}
\label{sec:rel_work_saliency}

Saliency maps provide a method for visualizing the representations learned by deep neural networks. The resulting map reveals which pixels will most impact the loss function when changed. Various methods for image-specific saliency maps have been proposed~\citep{springenberg2015striving,shrikumar2017just, pmlr-v70-sundararajan17a, Selvaraju_2017_ICCV, smilkov2017smoothgrad}. In this paper we utilize \citet{simonyan14}'s gradient based method. Recent independent evaluations~\citep{adebayo2020sanity, arun2020assessing} suggest this method is better than alternatives at exposing differences between a properly trained neural network and one with some randomized internal weights.

\subsection{Similarity Metrics}
\label{sec:rel_work_metrics}

Previous works have used several metrics for quantitative comparison between saliency maps, such as Spearman rank correlation with and without absolute value, the structural similarity index, and the Pearson correlation of the histogram of gradients \cite{adebayo2020sanity, arun2020assessing}. These metrics were found to be ill-suited to the comparison between saliency maps and reconstruction error maps. Although intense values are most important in both maps, directly comparing values across maps is impractical.

\section{Method}

In our formal setup, an input is a vector $x \in \mathbb{R}^n$. A deep autoencoder is a neural network model with an encoder $f(x): \mathbb{R}^n \rightarrow \mathbb{R}^m$ and decoder $g(z): \mathbb{R}^m \rightarrow \mathbb{R}^n$.

The \emph{reconstruction error map} $r(x) \in \mathbb{R}^n$ for image $x$ is defined as
\begin{equation}
r(x) = (x - g(f(x)))^2,
\end{equation}
producing a per pixel squared error map. The error at pixel index $i$ is denoted $r_i(x)$ The reconstruction loss $\bar{r}(x)$ is
\begin{equation}
\bar{r}(x) = \frac{1}{n} \sum_{i=1}^n r_i(x),
\end{equation}
the mean of $r(x)$ across pixel values. 

Following \citet{simonyan14}, a \emph{saliency map} $s(x) \in \mathbb{R}^n$ is defined as
\begin{equation}
s(x) = \Big| \frac{\delta \bar{r}}{\delta x} \Big|,
\end{equation}
which is found by backpropagating $\bar{r}(x)$.
In the case of multichannel images, the maximum of the channel values is taken for each pixel \cite{simonyan14}.

To evaluate whether reconstruction error maps correlate with important pixels in the input image, we compare $r(x)$ and $s(x)$. A saliency map $s(x)$ reveals which input pixels most impact $\bar{r}(x)$ when their values are changed. This differs from the $g(f(x))$ term of $r(x)$, where each output pixel no longer has a clear correspondence to an input pixel due to the transformations applied. Via convolutions and compression to an $m$-dimensional encoding vector, it is likely each input pixel has some impact on each pixel of $r(x)$.

If high values in $r(x)$ do indicate which parts of the input image are novel, a correspondence should exist between high values in $r(x)$ and high values in $s(x)$, where changes to the input $x$ have the greatest effect.

\subsection{Evaluation Metrics}

When comparing the reconstruction map $r(x)$ to the saliency map $s(x)$, it is important to note the maps are nonnegative but have different magnitudes. To facilitate comparison, both are scaled to $[0, 1]$ on a per image basis by dividing by the maximum pixel value for that map.

Visualizations of the maps are used to subjectively evaluate similarities. The saliency maps in the third column of Figure 1 are significantly more noisy than the reconstruction error maps in the last column. This is in part due to values in the saliency map being linearly related while the values in the reconstruction error map are quadratic. Therefore the square of the scaled map $s(x)$ is also presented for comparison between quadratic values.

To assess quantitative performance, we develop two metrics that measure \emph{correspondence} between the top K elements of two maps, as detailed below. These metrics require as input the top K largest elements (pixels) of each map, which are stored as a set.

\textbf{Metric 1: Top K agreement.}
We compute the \emph{top K agreement} between two maps as the size of the intersection of their selected sets. Top K agreement has similar goals as the Spearman rank discussed in Sec.~\ref{sec:rel_work_metrics} but provides clear two advantages for this problem. First, the method only considers the top K elements instead of the entire image, mitigating the impact of points with little importance. Second, this method depends only on the ordering of values and does not compare values across maps. Therefore this method produces the same output under any transformation of its inputs that maintains the same sorted order of pixel values. Thus transformations like scaling and squaring are not necessary to make saliency maps and reconstruction error maps comparable.

\textbf{Metric 2: Maximum distance to best match.}
Given the top K pixel sets of two maps, we compute the \emph{maximum distance to best match} by identifying for each member of the reference set the closest matching pixel (by Euclidean distance) in the other set. As the reference set, we select the \emph{saliency map}, as it is a more reliable indicator of which pixels should be identified as novel. We summarize all K matches via the maximum of such distances. This metric is motivated by a key limitation of top K agreement: it requires an exact pixel-level match and thus a shift of even one pixel reduces agreement substantially. Our distance metric provides a more gradual assessment developed for the intended use case of novelty localization, where a predicted map with small distances to the reference (ground truth) likely has milder consequences than one with large distances.    

\subsection{Experiment Setup}

Experiments are conducted on the MNIST \cite{mnist} dataset. The training set is split with 54,000 images for the training set and 6,000 images for a validation set. Digits 0, 1, 2, 3, and 4 are used as the normal classes while digits 5, 6, 7, 8, and 9 are used as novel classes. We use the convolutional autoencoder architecture for MNIST from Abati et al. \cite{abati19}, omitting their additional latent space model. The model is trained on the normal classes from the training set. The normal classes for the validation set were used to tune learning rate. Results were produced from the test set.

\section{Results}

We summarize our major findings in the paragraphs below.

\textbf{Finding 1: Reconstruction error maps often miss key salient input pixels.}
Visualization of the maps produced for randomly selected normal and novel inputs in Fig.~\ref{fig:visual_comparison_saliency_vs_reconstruction} reveals significant differences between saliency maps and reconstruction error maps. Saliency maps have more pixels with large values, revealing many pixels with high impact on the loss that are not found by the reconstruction error map. While using \emph{squared} saliency maps reduces the number of pixels with middling values, the reconstruction error still only captures a subset of the intense pixels. This suggests that reconstruction quality is sensitive to a larger number of input pixels than the reconstruction error map indicates.

\textbf{Finding 2: The correspondence between reconstruction error and saliency maps is weaker for novel images.}
The reconstruction error maps in Fig.~\ref{fig:visual_comparison_saliency_vs_reconstruction} often resemble an outline of the digit, perhaps most clearly with the “0” in row 3. This outline can also often be seen in the saliency maps for normal data. In contrast, both the outlined digit and its correspondence in the saliency maps is less visible in all 3 novel examples in Fig.~\ref{fig:visual_comparison_saliency_vs_reconstruction}. This suggests that the strength of correspondence between reconstruction and saliency maps depends on whether data is normal or novel.

\begin{figure}[t]
  \centering
  \includegraphics[width=\linewidth]{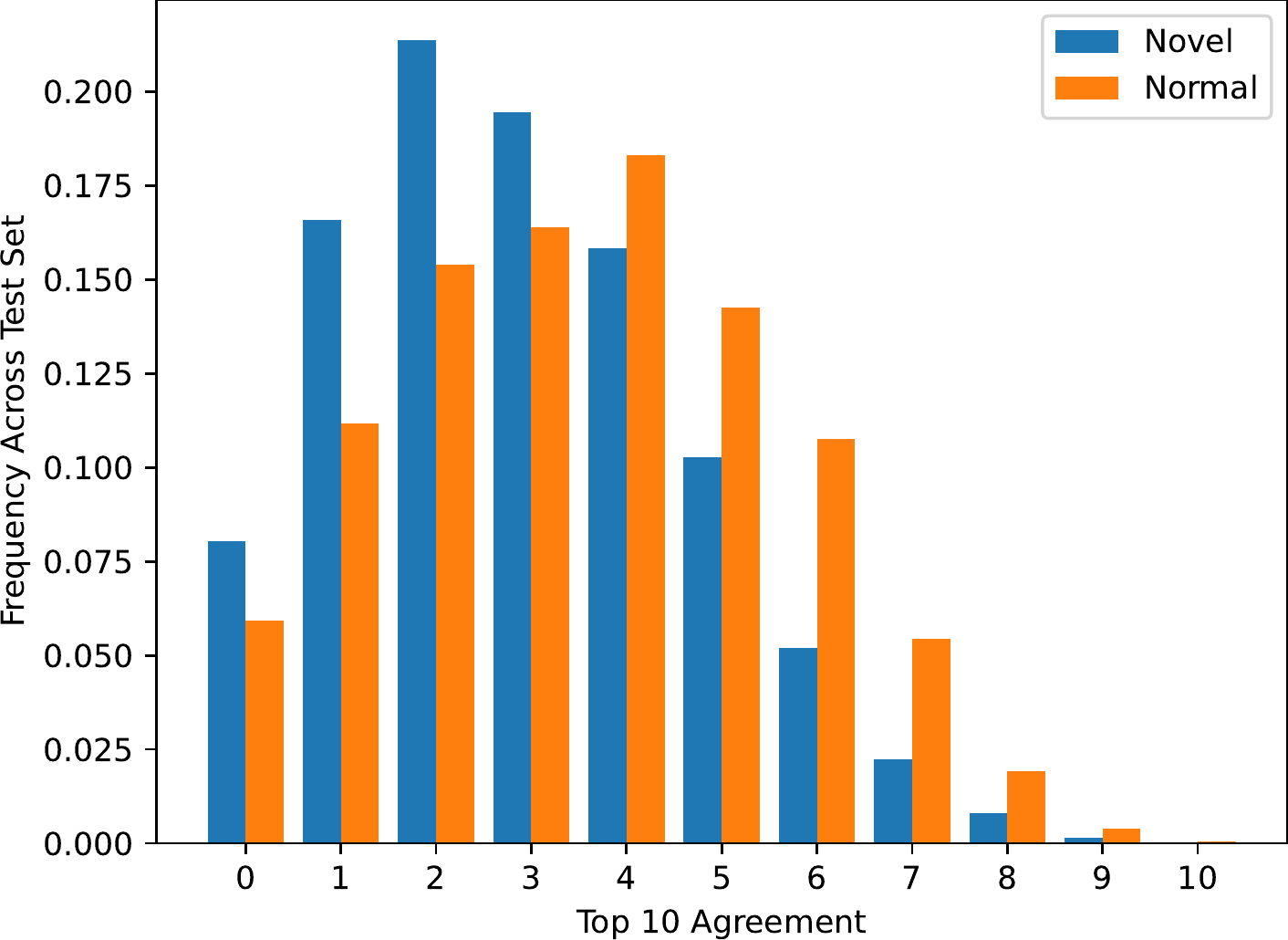}
\vspace{-18pt} 
\caption{The frequency of possible top 10 agreement values (0 = no agreement, 10 = a perfect match), observed across all images in the MNIST test set. Overall low agreement suggests that the top values of reconstruction error maps do not closely correlate with those of saliency maps, especially for novel images.}
	\label{fig:histogram_top_5_agreement_normal_vs_novel}
\end{figure}

\textbf{Finding 3: Top K agreement suggests reconstruction error does not closely correlate with saliency.}
Fig.~\ref{fig:histogram_top_5_agreement_normal_vs_novel} shows the frequency of top 10 agreement values across images in the test set. Total agreement is very rare; the most frequent level of agreement is 4 of 10 pixels for normal images and 2 of 10 pixels for novel images.

\textbf{Finding 4: Novel images have noticeably lower top K agreement than normal images.}
Tab.~\ref{tab:top_k_agreement} shows that only 26\% of normal images have the majority of their top 5 pixels agree across the two maps. This drops to 19\% for novel images. An even larger decrease is seen for top 10 agreement. Given the thousands of images in the test set, we expect this difference is significant.

\textbf{Finding 5: Maximum distance metric reveals the best match for at least one salient pixel is poor.}
For top 5 agreement we find that across all images in the test set, the mean max distance to best match is 8.28 pixels for normal images and 9.25 pixels for novel images. When expanding to the top 10 agreement, these numbers become 8.14 pixels for normal images and 8.90 pixels for novel images. As MNIST images are 28 by 28 pixels, these distances indicate a clear separation between the top K sets of the two maps.

\textbf{Finding 6: Use squared saliency maps when reconstruction maps use squared error.}
We find that both visually and quantitatively, squared saliency maps have better correspondence to reconstruction error maps than original (non-squared) saliency maps. When switching from saliency to squared saliency, the average mean-squared-error to reconstruction maps drops from 0.0303 to 0.0071 for normal images and from 0.0305 to 0.0072 for novel ones. When the reconstructions are based on squared error, it makes sense that a corresponding square is applied to the saliency map.

\begin{table}[t]
\vskip 0.15in
\begin{center}
\begin{small}
\begin{sc}
\begin{tabular}{lll}
\toprule
Top 5 Agreement & Normal Data & Novel Data \\
\midrule
0-2                   & 0.74        & 0.81       \\
3-5                   & 0.26        & 0.19 \\
\midrule
Top 10 Agreement & Normal Data & Novel Data \\
\midrule
0-5                    & 0.81        & 0.92       \\
6-10                   & 0.19        & 0.08       \\
\bottomrule
\end{tabular}
\end{sc}
\end{small}
\end{center}
\vskip -0.1in
\caption{The frequency of top 5 agreement and top 10 agreement values across normal and novel images in the MNIST test set. Frequencies are aggregated to show accuracy at identifying the majority of important input pixels. Both normal and novel images fail to exceed 26\% accuracy, with novel images significantly worse at top 10 agreement with only 8\% accuracy.}
\label{tab:top_k_agreement}
\end{table}

\section{Discussion \& Conclusion}

Our results find that reconstruction error maps do not closely correlate with the importance of pixels in the input images. Visually the maps are significantly different, although scaling the saliency map by squaring slightly improves this. The significant reduction of the mean squared error suggests squared saliency maps should be used over saliency maps when using techniques for image similarity sensitive to the scaling of pixel values.

The top K agreement clearly shows that reconstruction error maps are not identifying important pixels in the input, especially in the case of novel data. This suggests that while reconstruction error maps provide visually appealing novelty localization maps, they are not always representative of which parts of the input are novel and contribute most to reconstruction error. Further work and more experiments are needed to evaluate whether saliency maps themselves might provide a suitable replacement for novelty detection and localization.


\bibliography{autoencoder_saliency}
\bibliographystyle{icml2021}

%
%
%

\end{document}